\documentclass[11pt]{article}

\usepackage[utf8]{inputenc}
\usepackage[T1]{fontenc}
\usepackage[margin=1in]{geometry}
\usepackage{amsmath}
\usepackage{amssymb}
\usepackage{graphicx}
\usepackage{booktabs}
\usepackage{xcolor}
\usepackage{hyperref}
\hypersetup{colorlinks=true, linkcolor=black, citecolor=blue!45!black, urlcolor=blue!45!black}
\usepackage{xurl}
\usepackage[backend=biber,style=numeric,sorting=none]{biblatex}
\usepackage{authblk}
\usepackage{makecell}

\title{Predicting build orientation for SLM dental parts: a comparison of rotation representations and direct vector regression}

\author[1]{Felix Schmalzel\thanks{\texttt{felix.schmalzel@tha.de}}}
\author[2]{Reimar Waitz\thanks{\texttt{reimar.waitz@gmail.com}}}
\author[1]{Moritz Kronberger\thanks{\texttt{moritz.kronberger@tha.de}}}
\author[1]{Thorsten Sch\"oler\thanks{\texttt{thorsten.schoeler@tha.de}}}

\affil[1]{Technical University of Applied Sciences Augsburg, Augsburg, 86161, Germany}
\affil[2]{R. Waitz Data \& Science, Kaufering, 86916, Germany}

\date{}

\begin{document}

\maketitle

\begin{abstract}
Build orientation for selective laser melting (SLM) manufacturing of dental parts is usually chosen manually by technicians. We treat orientation prediction as supervised machine learning of the part's up-axis from technician-labeled production data, and test which rotation representations produce the best results. Using $n\approx2400$ patient-specific dental parts, we trained a ResNet-50 multi-view image backbone and a PointNeXt-S point-cloud backbone, both pretrained and fine-tuned end-to-end, on 13 up-axis representations spanning six classical $SO(3)$ parameterizations and seven representations defined directly on the unit sphere $S^2$.
We report the geodesic angular error between predicted and ground-truth up-axis on a test set, with and without test-time augmentation (TTA) over $K=21$ known rotations.
With TTA, the octahedral map achieves the lowest mean angular error ($10.6^\circ$, ResNet-50). The three lowest-error results overall are direct $S^2$ representations, though this may reflect label noise in the unsupervised in-plane component of the $SO(3)$ targets rather than a topological advantage. von Mises--Fisher collapses to a near-constant prediction when trained with PointNeXt-S but not with ResNet-50. TTA reduces mean angular error by 31--73\,\% across almost every representation and backbone. Overall, test-time augmentation over a small set of known rotations is the most consistent driver of accuracy, whereas the best-performing representation is strongly backbone-dependent.
\end{abstract}


\section{Introduction}
\label{sec:intro}

Selective laser melting (SLM) is a powder-bed fusion process in which a laser fuses metal powder layer-by-layer to build up a part. In dental laboratories it is widely used to produce crowns, bridges, partial denture frameworks, and implant abutments directly from digital designs \cite{ref1,ref2,ref3,ref4}. Because parts are built layer-by-layer, the orientation in which a part is placed in space (its \emph{build orientation}) has a strong effect on surface quality, dimensional accuracy, support requirements, post-processing effort, and process time \cite{ref5,ref6,ref7}. Choosing this orientation is typically done manually by experienced technicians. Automating it is attractive because it removes a repetitive step from the digital workflow and reduces the dependence on scarce expertise.

Uprighting of dental parts is not well researched. Uprighting in general has received attention, both as an optimization problem \cite{ref8,ref9,ref10} and, more recently, through learning-based methods \cite{ref11,ref12}, but existing methods mostly target parts with flat surfaces and a stable connection to the printer's build plate, neither of which applies to dental parts.
SLM dental printing imposes a different set of constraints: downward-facing skin is of lower quality and must be connected to supports \cite{ref13,ref14,ref15}, while anatomical faces need to point upward to preserve accuracy \cite{ref16,ref17}. Parts are generally thin-walled and small \cite{ref18,ref19}, and supports cannot be placed arbitrarily on anatomical surfaces \cite{ref20,ref7}. Together, these constraints mean that build directions derived for classical CAD parts do not transfer to the dental setting.

When finding build orientations, technicians balance several objectives simultaneously: minimizing support contact on anatomically critical faces, maintaining support accessibility for removal, respecting overhang limits, controlling thermal distortion, and preserving aesthetic fit \cite{ref21,ref22,ref23}. The trade-offs between these objectives depend on the specific geometry of each patient-specific part. Much of the relevant knowledge is implicit: experienced operators rely on pattern recognition built up over thousands of cases rather than on an explicit rule set \cite{ref24}. Attempts to encode this as a fixed multi-objective cost function miss this implicit expert knowledge. We therefore treat uprighting as a supervised learning problem, using orientations produced by professional technicians as ground truth, in line with recent learning-based approaches to build-orientation prediction \cite{ref25,ref26,ref27}. The closest prior work in the dental setting is Lin et al.~\cite{ref28}, who target partial fixed dental prostheses with a hybrid pipeline: PointNet++ classifies the prosthesis as single- or multi-unit, a second network segments the seating-hole region, and NSGA-II then searches an explicit support-area / build-height objective constrained to $\pm45^\circ$ around a fixed z-axis, returning a Pareto set from which the operator selects. Their approach depends on a hand-crafted multi-objective cost and on annotating a single localized critical region per class, and presupposes that the build direction lies near a known reference. In contrast, we regress the full $S^2$ up-axis end-to-end from technician-labeled orientations, without an explicit cost function or a localized reference region, which extends more readily to part classes such as partial denture frameworks, where the geometric features that matter for orientation are distributed across the part rather than concentrated in one hole.

The choice of rotation parameterization directly influences neural network training behavior. Any continuous mapping from an input to a rotation must traverse the topology of $SO(3)$ (or, in our case, $S^2$), and common parameterizations (Euler angles, quaternions, axis--angle) are either discontinuous, double-covering, or both. Such discontinuities create large gradients near singularities and degrade the trainability of neural networks \cite{ref29,ref30,ref31}. Overparameterized representations, such as the first two columns of a rotation matrix followed by Gram--Schmidt orthogonalization, avoid these pathologies and empirically converge to lower error \cite{ref29,ref32,ref33}. For our use case the rotation around the z-axis is irrelevant, reducing the prediction target to $S^2$ and opening up additional representations, including direct Cartesian regression and parameterized distributions on the sphere \cite{ref34,ref35,ref36}. Inspired by prior work on $SO(3)$ representations \cite{ref29,ref31}, we discuss, both theoretically and experimentally, which representation performs best for $S^2$ prediction in this setting.

We compare six classical $SO(3)$ parameterizations against seven representations defined directly on $S^2$ (Section~\ref{sec:representations}), evaluating all of them with a ResNet-50 \cite{ref40} multi-view image backbone and a PointNeXt-S \cite{ref51} point-cloud backbone.
Our main metric is the angle between the ground-truth up-axis and the predicted up-axis, which is invariant to rotation around the z-axis \cite{ref42}.

\section{Materials and Methods}
\label{sec:methods}

\subsection{Dataset}
\label{sec:dataset}

We use $n \approx 2400$ patient-specific dental parts. All parts originate from production batches that were physically manufactured. Every part therefore carries an orientation chosen by an experienced technician for that print, which we treat as the ground truth. Geometry files are fully anonymized and contain no patient-identifiable information. The dataset is split into training, validation, and test sets at a 0.8/0.1/0.1 ratio. Splitting is performed within each production batch rather than at the batch level, so every batch contributes parts to all three splits in the same proportion. The dataset does not include class labels (e.g. crown, bridge, framework, abutment) for the individual parts, so all experiments are conducted on the dataset as a whole without class-conditional analysis.

\subsection{Preprocessing}
\label{sec:preprocessing}

Each mesh is normalized independently by centering and scaling it to a unit sphere. This scale factor is computed per part rather than from a fixed, dataset-wide constant, so absolute part size is discarded and only shape is available to the model as a cue. No rotation is applied at this stage to preserve the technician-chosen orientation as the regression target.

\subsection{Input modalities}
\label{sec:input-modalities}

We evaluate two input modalities, one image-based (ResNet-50) and one point-based (PointNeXt-S).

\paragraph{Multi-view images.} For each part we render $n_{\text{view}} = 3$ orthographic views at a resolution of $224 \times 224$ pixels. Rather than using conventional rendering we use a custom ray-based renderer, which encodes different information in the color channels:

\begin{itemize}
    \item \emph{X-ray channel}: the sum of signed entry/exit depth differences along the ray, encoding the cumulative thickness of solid material the ray passes through.
    \item \emph{Surface channel}: the depth of the first ray--mesh intersection, encoding the visible front surface.
    \item \emph{Layer-count channel}: a logarithmic encoding of the number of intersections, $\log(c/16 + 1)$, capturing internal geometric complexity.
\end{itemize}
\noindent
The resulting tensor per part has shape $(n_{\text{view}}, 3, 224, 224)$.
The $n_{\text{view}}$ camera positions are distributed on the unit sphere using a Fibonacci lattice \cite{ref43,ref44}, which provides an approximately uniform coverage of $S^2$ for any $n_{\text{view}}$ and therefore lets us ablate the number of views without redesigning the camera set, unlike an axis-aligned cube-face arrangement $\{\pm e_x, \pm e_y, \pm e_z\}$ that would be tied to $n_{\text{view}} = 6$.

\paragraph{Point clouds.} For each part we sample 1024 points by farthest-point sampling on the normalized vertex set. Each input is therefore a tensor of shape $(1024, 3)$ containing only the $(x, y, z)$ coordinates. No surface normals or other features are attached. If a mesh has fewer than 1024 vertices, the missing points are filled by sampling with replacement from the existing vertices. Farthest-point sampling is preferred over uniform random sampling because it preserves geometric coverage of the part's surface and the location of small features, both of which carry information about the technician-chosen orientation.

\subsection{Rotation and up-axis representations}
\label{sec:representations}

We compare classical $SO(3)$ parameterizations applied to the full rotation against parameterizations defined directly on $S^2$ that target only the up-axis. For the six $SO(3)$ representations, a full rotation target is needed even though only the up-axis is supervised. We obtain one by Gram--Schmidt completion of the ground-truth up vector against a fixed reference axis. The in-plane (z-axis) rotation this introduces is arbitrary and is never supervised or evaluated. Only the third column of the decoded rotation matrix, i.e.\ the up-axis, is used for loss and metrics computation.

\paragraph{$SO(3)$ representations.}
\emph{Quaternion}: the network outputs an unnormalized 4-vector, renormalized to a unit quaternion; the up-axis is the third column of the corresponding rotation matrix. The loss is $1-\langle q_{\text{pred}},q_{\text{gt}}\rangle^2$, which makes the loss \emph{value} invariant to the double-cover sign ambiguity, but $q$ and $-q$ remain two distinct global minima in the 4D output space, so the network still has to commit to one antipodal solution rather than being given a single unambiguous target.
\emph{Rotation matrix}: 9 outputs, reshaped to a $3\times3$ matrix and projected onto $SO(3)$ by its nearest orthogonal matrix (SVD); loss is the geodesic distance to the ground-truth matrix.
\emph{Rotation matrix with Gram--Schmidt orthogonalization} (``6D''): 6 outputs, interpreted as the first two columns of a rotation matrix and completed to $SO(3)$ by Gram--Schmidt orthogonalization \cite{ref29} instead of SVD projection.
\emph{Euler angles}: three angles under a fixed $Z$-$Y$-$X$ convention, regressed with mean squared error (MSE) on the angles.
\emph{Axis--angle}: a unit axis and a scalar angle $\theta\in[0,\pi]$ regressed jointly (4 outputs), with a geodesic loss on the implied rotation.
\emph{Exponential coordinates}: the 3-vector $\theta\hat n$ itself\footnote{The axis--angle pair ($\hat{n}$, $\theta$) encoded as a single 3-vector $\theta\hat{n}$ (also known as the rotation vector).}, decoded through the matrix exponential, with a geodesic loss. For all six, only the decoded up-axis is evaluated.

\begin{figure}[htbp]
    \centering
    \includegraphics[width=\textwidth]{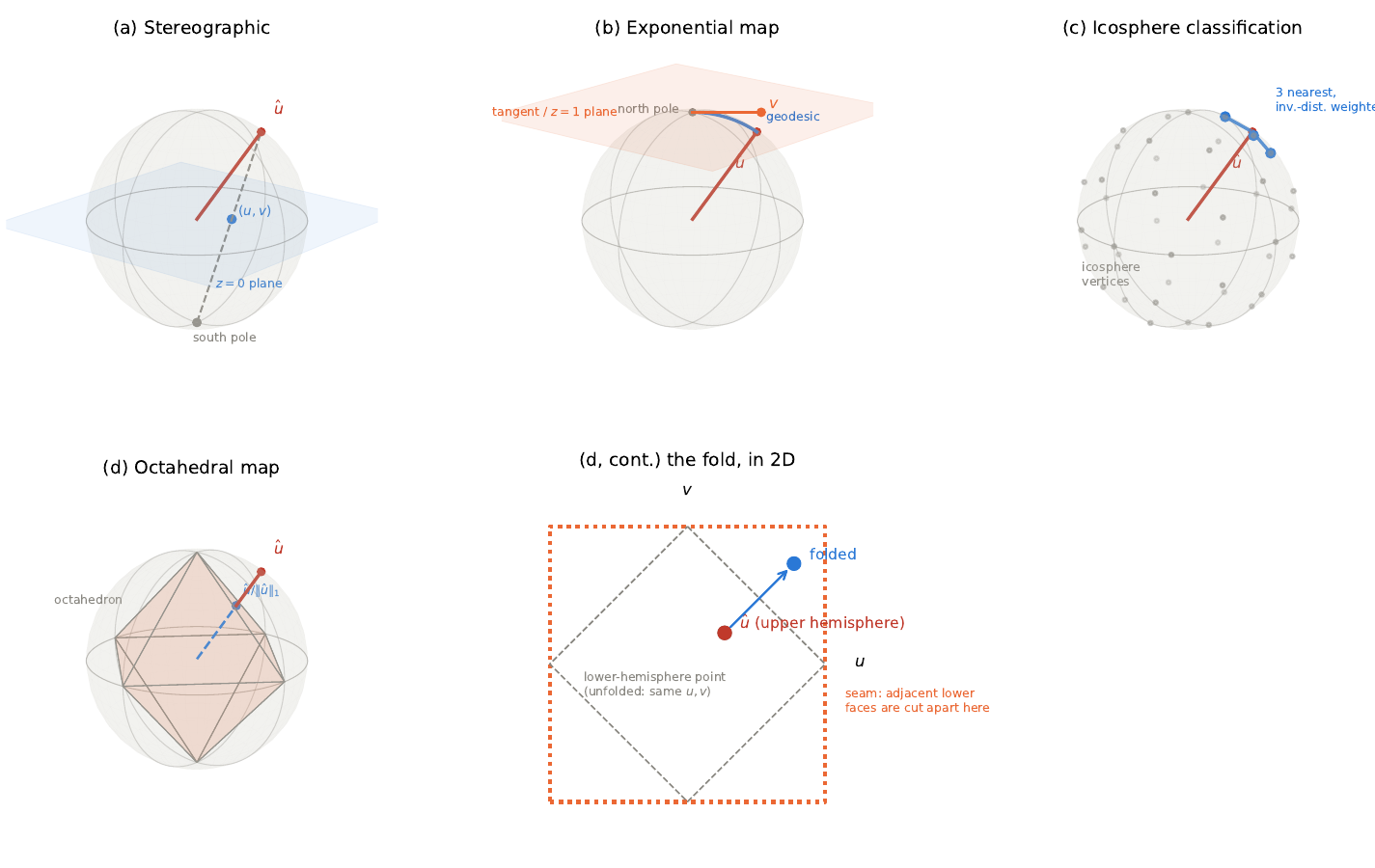}
    \caption{The four representations that are hardest to picture from the description above, all shown for the same example up-axis $\hat u$ (red). (a) Stereographic: projection from the south pole onto the plane $z=0$. (b) Exponential map: the tangent-plane vector $v$ at the north pole and its corresponding geodesic (blue) on the sphere. (c) Icosphere classification: the discretized vertex set, with the three nearest vertices (blue) weighted by inverse distance. (d) Octahedral map: radial projection onto the unit-$L_1$ octahedron (left), and why the fold is needed (right): without it, $\hat u$ and the marked point on the opposite (lower) hemisphere would land on the identical $(u,v)$. Folding the lower hemisphere into the square's free corners removes the ambiguity.}
    \label{fig:representations-detail}
\end{figure}

\paragraph{$S^2$ representations (Figure~\ref{fig:representations-detail}).}
\emph{Cartesian}: a raw 3-vector, $L_2$-normalized to $\hat u\in S^2$; loss $1-\hat u_{\text{pred}}\cdot\hat u_{\text{gt}}$.
\emph{Spherical}: the polar/azimuthal pair $(\theta,\varphi)$; the pair is decoded to a direction and the loss is the geodesic distance to the ground-truth direction rather than MSE on $(\theta,\varphi)$ directly, since $\varphi$ wraps at $\pm\pi$.
\emph{Stereographic}: projection of $S^2\setminus\{\text{south pole}\}$ onto $\mathbb{R}^2$ from the south pole, $(u,v)=\bigl(x/(1+z),\,y/(1+z)\bigr)$, so the singularity falls opposite the upward region where targets concentrate; as with spherical coordinates, the $(u,v)$ pair is decoded to a direction and the loss is $1-\hat u_{\text{pred}}\cdot\hat u_{\text{gt}}$ on the decoded direction rather than MSE in $\mathbb{R}^2$.
\emph{Exponential map}: the tangent-space exponential map at the north pole $p=(0,0,1)$, $\mathrm{Exp}_p(v)=\cos\lVert v\rVert\,p+\mathrm{sinc}\lVert v\rVert\,v$, regressing the 2D tangent vector $v$ with MSE against $\mathrm{Log}_p(\hat u_{\text{gt}})$.
\emph{Octahedral map}: $\hat u$ is $L_1$-normalized onto the octahedron and the lower hemisphere folded into the $[-1,1]^2$ square, giving a bijective 2D encoding regressed with MSE after a $\tanh$ output nonlinearity.
\emph{Icosphere classification}: $S^2$ is discretized into the $V$ vertices of a subdivided icosphere ($\nu=20\Rightarrow V=4002$); the network outputs $V$ logits, trained with cross-entropy against the single nearest vertex to the target and decoded as the softmax-weighted directional mean of the vertices, the only classification-based formalism. The formalism separately defines an inverse-distance-weighted soft target over the three nearest vertices to the ground-truth direction, used only for round-trip encode/decode checks, not for the training loss above.
\emph{von Mises--Fisher (vMF)}: a 4D output $(\mu_{\text{raw}},\log\kappa)$ parameterizes a unimodal directional distribution with mean $\mu=\mathrm{normalize}(\mu_{\text{raw}})$ and concentration $\kappa=\mathrm{softplus}(\log\kappa)$ \cite{ref35,ref37}, trained by maximizing the vMF log-likelihood; $\mu$ is the point prediction and $\kappa$ doubles as an uncertainty estimate.

\subsection{Backbones and prediction head}
\label{sec:backbones}

We use ResNet-50 \cite{ref40} as the image backbone and PointNeXt-S \cite{ref51} as the point-cloud backbone, both initialized from pretrained weights despite the substantial domain gap to our renders and meshes: ResNet-50 from the \texttt{a1} ImageNet-1k \cite{ref45,ref46} recipe distributed via \texttt{timm}, and PointNeXt-S from the official checkpoint trained on ModelNet40 \cite{ref52}. Pretrained features transfer even across large domain shifts \cite{ref50}, and empirically we found fine-tuning from these checkpoints to converge faster and to lower error than training either backbone from scratch. Both backbones are fine-tuned end-to-end with gradual unfreezing (Section~\ref{sec:training}) rather than kept frozen, and both feed the same two-layer multilayer perceptron (MLP) head (hidden dimension 512, ReLU, dropout 0.2) whose output dimension $k$ is the parameter count of the chosen representation. Only $k$ varies across backbone--representation combinations.

\subsection{Training}
\label{sec:training}

All models are trained end-to-end with AdamW \cite{ref48} (weight decay $10^{-4}$), with separate learning rates for backbone ($10^{-5}$) and head ($10^{-3}$) on a cosine annealing schedule \cite{ref49} over up to $E_{\max}=120$ epochs, with early stopping (patience 20 epochs on validation mean angular error) and gradient-norm clipping at 1.0. Because both backbones start from pretrained weights (Section~\ref{sec:backbones}), we unfreeze them gradually rather than fine-tuning every parameter from step one: the backbone starts fully frozen, and one additional stage is unfrozen at each of epochs 5, 10, 15, 20, and 25, latest layers first (e.g.\ for ResNet-50, \texttt{layer4} at epoch 5 down to the stem at epoch 25), so early training only adapts the randomly initialized head before backbone features are perturbed.

We apply rotation augmentation to the training set: at each training step the input mesh is rotated by a different random rotation drawn uniformly on $SO(3)$, and the regression target is rotated accordingly so that the technician-chosen orientation moves with the part. Every raw mesh is stored in the pose it was actually printed in, so evaluating validation and test parts directly in that stored pose would let a model exploit incidental artifacts of how the technician-chosen orientation happens to sit in the file's own coordinate frame. Section~\ref{sec:evaluation} describes the fixed, known rotations applied to validation and test parts to limit this.

\subsection{Evaluation}
\label{sec:evaluation}

Our primary metric is the geodesic angle on $S^2$ between the predicted and ground-truth up-axis, which is invariant to rotation around the build z-axis \cite{ref42}. We report the mean and median angular error on the test set, together with the cumulative distribution of errors at fixed thresholds.

Both validation and test parts are evaluated under a fixed, shared set of $K=21$ known rotations applied to each part's raw, as-printed mesh, rather than under a single per-sample random rotation: we take the $K=21$ directions on the upper hemisphere of a twice-subdivided icosphere and, for each, build the rotation matrix that carries the north pole onto that direction. One of these 21 directions is the north pole itself, so one of the 21 views is the identity rotation and recovers the stored pose exactly. For each part we obtain $K$ predictions under these $K$ known rotations: for the point-cloud backbone the rotation is applied to the input point cloud, while for the image backbone the rotation is applied to the mesh and the $n_{\text{view}}$ Fibonacci views are re-rendered. Because each rotation is known, every per-view prediction can be transformed back into the part's raw, as-printed (canonical) frame.

How the $K$ per-view predictions are combined depends on the representation. For most representations, each of the $K$ predictions are decoded to a unit up-axis, every direction is unrotated into the canonical frame using its known rotation, then the $K$ canonical directions are averaged and renormalized to $S^2$.
Two representations depart from this. For the von Mises--Fisher formalism we combine the $K$ predicted distributions via the product of von Mises--Fisher densities, i.e.\ a concentration-weighted sum of the unrotated mean directions, so confident (high-$\kappa$) views are given proportionally more weight than uncertain ones rather than every view counting equally. For the icosphere classification formalism, predictions live in per-vertex probability space rather than on $S^2$ directly and vertex indices are frame-dependent, so they cannot be unrotated the way a continuous direction can. Instead, for each rotation we remap its softmax distribution onto a shared canonical vertex indexing (found by nearest-neighbor matching of each rotated vertex to its unrotated counterpart), average the $K$ remapped distributions, and decode the result once.

We report two numbers from this same set of $K$ predictions: \emph{No-TTA} is the mean error of the $K$ individual, un-aggregated per-view predictions, and \emph{TTA} is the error of the single prediction obtained by aggregating across the same $K$ views (above). Both are therefore computed under this fixed, known set of $K$ reorientations rather than under one arbitrary unknown rotation. Reporting both separates the contribution of view-aggregation itself from the architectural effects of the backbone and the rotation representation.

\section{Results}
\label{sec:results}

\begin{figure}[htbp]
    \centering
    \includegraphics[width=0.85\textwidth]{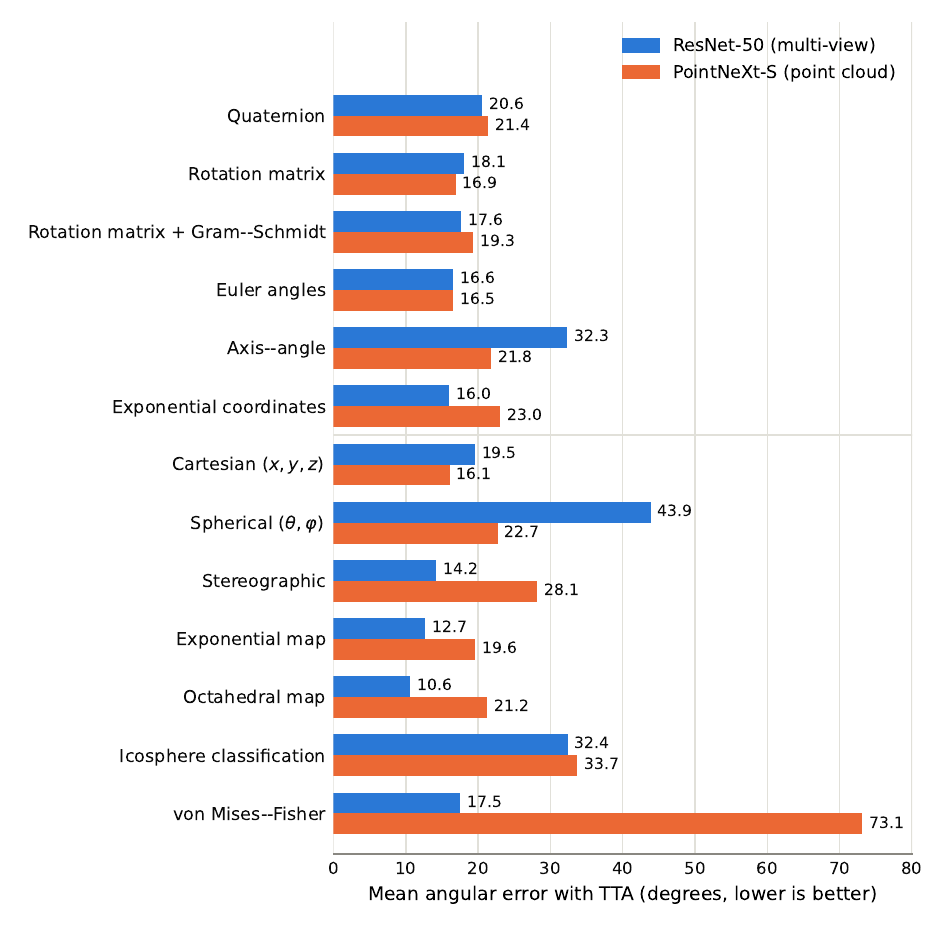}
    \caption{Mean angular error with TTA (test set, $K=21$ icosphere rotations), by representation and backbone. Lower is better. The top six representations parameterize the full rotation ($SO(3)$); the bottom seven target the up-axis directly ($S^2$).}
    \label{fig:results-bar}
\end{figure}

For each of the 13 representations and both backbones, Figure~\ref{fig:results-bar} and Table~\ref{tab:results} report the mean and median geodesic error between the predicted and ground-truth up-axis on the held-out test set together with the fraction of test parts predicted within $10^\circ$ of the ground truth, with and without test-time augmentation.

\begin{table}[htbp]
    \centering
    \caption{Test-set angular error (degrees) and fraction of parts within $10^\circ$, without and with TTA, for each representation and backbone. Bold marks the best mean-with-TTA error for each backbone and parameterization paradigm ($SO(3)$ vs. $S^2$).}
    \label{tab:results}
    \small
    \setlength{\tabcolsep}{4pt}
    \begin{tabular}{lcccccccc}
        \toprule
        & \multicolumn{4}{c}{ResNet-50} & \multicolumn{4}{c}{PointNeXt-S} \\
        \cmidrule(lr){2-5} \cmidrule(lr){6-9}
        Representation & \makecell{No-TTA\\mean} & \makecell{TTA\\mean} & \makecell{TTA\\med.} & \makecell{TTA\\\%$<10^\circ$} & \makecell{No-TTA\\mean} & \makecell{TTA\\mean} & \makecell{TTA\\med.}\ & \makecell{TTA\\\%$<10^\circ$} \\
        \midrule
        Quaternion & 37.0 & 20.6 & 11.4 & 44.7 & 37.5 & 21.4 & 14.8 & 24.1 \\
        Rotation matrix & 34.9 & 18.1 & 8.8 & 60.9 & 27.9 & 16.9 & 9.8 & 51.2 \\
        Rotation matrix + Gram--Schmidt & 38.7 & 17.6 & 11.0 & 45.3 & 43.5 & 19.3 & 12.7 & 38.1 \\
        Euler angles & 54.1 & 16.6 & 13.9 & 30.0 & 58.8 & \textbf{16.5} & 15.8 & 23.8 \\
        Axis--angle & 58.7 & 32.3 & 21.0 & 23.8 & 64.1 & 21.8 & 18.5 & 16.2 \\
        Exponential coordinates & 43.8 & \textbf{16.0} & 12.3 & 36.6 & 51.9 & 23.0 & 15.0 & 30.0 \\
        \midrule
        Cartesian $(x,y,z)$ & 35.2 & 19.5 & 9.6 & 52.8 & 26.2 & \textbf{16.1} & 10.2 & 49.1 \\
        Spherical $(\theta,\varphi)$ & 63.9 & 43.9 & 34.3 & 4.4 & 44.1 & 22.7 & 16.9 & 20.0 \\
        Stereographic & 39.8 & 14.2 & 9.8 & 50.9 & 46.0 & 28.1 & 19.5 & 23.4 \\
        Exponential map & 40.9 & 12.7 & 10.4 & 48.1 & 37.4 & 19.6 & 12.9 & 36.6 \\
        Octahedral map & 39.8 & \textbf{10.6} & 9.2 & 57.5 & 40.5 & 21.2 & 17.9 & 19.1 \\
        Icosphere classification & 58.7 & 32.4 & 30.9 & 0.0 & 62.9 & 33.7 & 30.8 & 2.8 \\
        von Mises--Fisher & 32.5 & 17.5 & 10.8 & 46.6 & 76.3 & 73.1 & 73.1 & 0.0 \\
        \bottomrule
    \end{tabular}
\end{table}

\paragraph{Representations.}
With TTA, the three lowest mean errors on ResNet-50 are the octahedral map ($10.6^\circ$), the exponential map ($12.7^\circ$), and the stereographic projection ($14.2^\circ$).
All three are two-dimensional representations defined directly in $S^2$.
On ResNet-50 all three beat every $SO(3)$ representation (lowest $SO(3)$ mean is $16.0^\circ$).
On PointNeXt-S the lowest mean errors come from a different set of representations: Cartesian ($16.1^\circ$), Euler angles ($16.5^\circ$), and the rotation matrix ($16.9^\circ$).
These rankings are also metric dependent, with ranks in mean angular error not being identical to ranks in within-$10^\circ$ fraction.


\paragraph{Failure modes.}
Two representations perform clearly worse than the rest.
Icosphere classification has the lowest within-$10^\circ$ fraction on both backbones ($0.0\,\%$ on ResNet-50 and $2.8\,\%$ on PointNeXt-S), meaning it rarely places the part within $10^\circ$ of the target.
Most strikingly, von Mises--Fisher with PointNeXt-S collapses to a degenerate fit: its test standard deviation is $0.07^\circ$ around a mean of $73.1^\circ$, i.e. the model converges to an almost constant prediction rather than one that varies with the input, while the same formalism trained normally on ResNet-50 ($17.5^\circ$ mean). While we were unable to identify the exact cause at the time of writing, we assume it is unrelated to the general PointNeXt-S vs.\ ResNet-50 comparison, since every other $S^2$ representation trained successfully on both backbones.


\paragraph{TTA.} Averaging over $K=21$ known rotations reduces the mean angular error substantially and uniformly: across every representation and backbone except the one degenerate vMF run, TTA cuts the mean error by 31--73\,\%, e.g.\ from $39.8^\circ$ to $10.6^\circ$ for the octahedral map on ResNet-50, and from $58.7^\circ$ to $32.3^\circ$ for axis--angle.

\paragraph{Backbones.}
ResNet-50 outperforms PointNeXt-S with TTA on 8 of 13 representations, most clearly for representations whose $S^2$ decoding is sensitive to output scale (stereographic, exponential map, and octahedral map).
Conversely, PointNeXt-S outperforms ResNet-50 most clearly on a mixed set of representations (spherical coordinates, axis--angle, and Cartesian coordinates).


\section{Discussion}
\label{sec:discussion}

The most clearly interpretable result is the effect of test-time augmentation.
Its uniform benefits across representations indicate that a large part of the No-TTA error reflects actual per-view sensitivity to viewing angle under the fixed set of known rotations (Section~\ref{sec:evaluation}), rather than an inherent ceiling on how well a representation can localize the up-axis once multiple views are combined.

The continuity argument from Section~\ref{sec:intro} states that discontinuous or double-covering representations are expected to train worse than overparameterized, continuous ones \cite{ref29,ref30,ref31}.
This is not clearly supported by experiment.
Euler angles, a discontinuous parameterization, are among the most competitive and backbone-stable $SO(3)$ representations on mean error.
Additionally, the 6D Gram--Schmidt representation was proposed specifically to remove the residual discontinuity of naive matrix regression \cite{ref29}, yet it is outperformed by the 9D rotation-matrix-via-SVD projection on PointNeXt-S (although the opposite is true on ResNet-50). This agrees with Levinson et al.'s later finding that SVD orthogonalization may outperform Gram--Schmidt in practice \cite{ref32} despite the latter's cleaner continuity argument.


A pattern the continuity argument does not address at all is that the overall top-3 lowest-error representations are all direct $S^2$ representations (octahedral map, exponential map, stereographic), ahead of every $SO(3)$ representation. We hypothesize this is primarily due to label noise in the training target and not topology: every $SO(3)$ representation in our setup regresses toward a full rotation matrix whose up-axis column is the technician-chosen orientation but whose other two columns are an arbitrary in-plane rotation that carries no signal about the task. This is fixed only by our Gram--Schmidt reference axis (Section~\ref{sec:representations}). Direct $S^2$ regression never constructs or supervises these unused rotation parameters, so all of the network's capacity is spent on the two degrees of freedom that are actually evaluated. This hypothesis is specific to our target-construction choice and not a general claim about $SO(3)$ representations. A controlled ablation, e.g.\ supervising the in-plane rotation with a real, part-specific reference instead of an arbitrary one, would be needed for confirmation.

The failure of icosphere classification can be explained from the implementation: the training loss uses plain cross-entropy against the single nearest vertex, so predicting an adjacent vertex a few degrees away is penalized identically to predicting the antipodal one. 
With $V\approx4000$ vertices, $n_\text{train}\approx1900$ training parts and $21 \leq m \leq 120$ epochs, each vertex was on average the nearest-neighbor target during training for $10 \leq \frac{n_\text{train} \cdot m}{V} \leq 57$ times. Both point to underused signal rather than a fundamental limitation of discretized $S^2$ classification. We would expect a distance-aware loss, e.g.\ a soft target over the nearest few vertices instead of a single hard one, to close much of the gap to the continuous representations.

von Mises--Fisher's \texttt{decode} is identical to Cartesian's ($L_2$-normalizing the raw output direction), so vMF is Cartesian regression with an additional scalar $\kappa$ and a different (negative log-likelihood) loss. This framing is consistent with our results: on ResNet-50, vMF performs comparably to Cartesian rather than clearly better ($17.5^\circ$ vs.\ $19.5^\circ$ mean, but worse median and $\%<10^\circ$; Table~\ref{tab:results}).
On PointNeXt-S, $\kappa$ is the only differentiating factor between vMF and Cartesian. Since every other $S^2$ representation trains successfully on that backbone, we suspect $\kappa$ as a plausible culprit for the collapse (Section~\ref{sec:results}), though we have not confirmed this directly.

On backbones, ResNet-50 outperforms on more representations, but PointNeXt-S is ahead for Cartesian coordinates, the rotation matrix, and axis--angle, and is the only backbone under which spherical coordinates reach a usable accuracy. Both backbones are pretrained far outside our domain (natural photographs for ResNet-50, rigid CAD objects from ModelNet40 for PointNeXt-S). We therefore cannot attribute the backbone gap to which pretraining domain is closer to dental parts. Additionally, input modality (rendered multi-view images vs.\ raw point coordinates) and the very different capacity/inductive-bias profile of the two architectures are confounded with the pretraining source in this comparison.

\paragraph{Limitations.}
This is a single-institution dataset and we report no class-conditional breakdown (crown vs.\ bridge vs.\ framework vs.\ abutment) because class labels are not available (Section~\ref{sec:dataset}). Orientation difficulty plausibly differs by class.
Additionally, several representation--backbone combinations rest on one training run rather than an average over independent seeds, so we cannot separate actual representation gaps from run-to-run variance for close calls (e.g.\ Euler vs.\ exponential coordinates on ResNet-50).

\section{Conclusions}
\label{sec:conclusions}

We treat build-orientation prediction for SLM dental parts as supervised machine learning of the up-axis on $S^2$ from technician-labeled orientations and compare 13 rotation and up-axis representations across two backbones on $n\approx2400$ patient-specific parts. 
Test-time augmentation over a small set of known rotations reduces mean angular error by 31--73\,\% across representations and backbones (excluding degenerate runs), confirming that single-view predictions are less reliable than aggregated ones.
The three lowest errors overall stem from low-dimensional $S^2$ representations on ResNet-50, led by the octahedral map ($10.6^\circ$ mean angular error).
However, this result may reflect label noise in the unsupervised in-plane component of $SO(3)$ rather than a topological advantage.
In general, the best-performing representations are strongly backbone-dependent, as ranking does not transfer between the two backbones.
Since they differ in input modality, architecture, and pretraining, this interaction cannot be attributed to any single factor.
These results are not predicted by the continuity argument (Section~\ref{sec:intro}) motivating the representation choices: discontinuous parameterization is not a reliable predictor of worse performance.
Two failure modes remain open for future research: icosphere classification's use of hard rather than soft targets, and von Mises--Fisher's training collapse on PointNeXt-S. Also open is a class-conditional analysis once part-type labels are available.


\section*{CRediT Author Contribution Statement}

\textbf{Felix Schmalzel}: conceptualization, formal analysis, investigation, methodology, software, validation, visualization, writing (original draft, review \& editing).
\textbf{Reimar Waitz}: conceptualization, methodology, writing (review \& editing).
\textbf{Moritz Kronberger}: writing (review \& editing), project administration.
\textbf{Thorsten Sch\"oler}: conceptualization, funding acquisition, project administration, supervision, writing (review \& editing).

\section*{Declaration of Competing Interest}

The authors declare that they have no known competing financial interests or personal relationships that could have appeared to influence the work reported in this paper.

\section*{Funding}

This work was supported by the Free State of Bavaria within the Bavarian Joint Research Program (BayVFP), funding line Digitalization, administered by VDI/VDE Innovation + Technik GmbH [grant numbers DIK-2309-0018, DIK0550/02].

\section*{Ethical Approval}

This study used fully anonymized CAD geometry files without any patient-identifiable information. No ethical approval was required.

\section*{Data Availability}

The dataset used in this study is not publicly available.

\printbibliography

\end{document}